\documentclass[11pt,a4paper]{article}
\usepackage[hyperref]{naaclhlt2018}
\usepackage{times}
\usepackage{latexsym}
\usepackage[utf8]{inputenc}
\usepackage{graphicx}
\usepackage[]{caption}
\DeclareCaptionFont{tenpt}{\fontsize{10pt}{10pt}\selectfont #1}
\usepackage[]{subcaption}
\usepackage{pgfplots}
\usepackage{multirow}
\usepackage{amsmath,enumitem}
\usepackage{amssymb,todonotes}
\usepackage{pgfplotstable}
\usepackage{url}
\usepackage{breakurl} 

\definecolor{c8}{RGB}{225,0,0}
\definecolor{c7}{RGB}{228,26,28}
\definecolor{c6}{RGB}{55,126,184}
\definecolor{c5}{RGB}{77,175,74}
\definecolor{c4}{RGB}{152,78,163}
\definecolor{c3}{RGB}{255,127,0}
\definecolor{c2}{RGB}{204, 204, 0}
\definecolor{c1}{RGB}{166,86,40}
\pgfplotscreateplotcyclelist{sf_colors}{%
        {c6, mark=square},%
        {c5, mark=x},%
        {c4, mark=+},%
        {c7, mark=*},%
     }   

\pgfplotscreateplotcyclelist{bar_colors}{%
 		{c6, fill=c6, mark=none},%
        {c5, fill=c5, mark=none},%
        {c4, fill=c4, mark=none},%
        {c3, fill=c3, mark=none},%
     } 
 
\pgfplotscreateplotcyclelist{centroid_colors}{%
 		{c6,  mark=none},%
        {c8,  mark=none},%
        {c6,  mark=none},%
        {c8,  mark=none},%
     } 
     
\pgfplotscreateplotcyclelist{centroid_colors1}{%
        {c8,  mark=none},%
     }

\aclfinalcopy 

\newcommand\blfootnote[1]{%
  \begingroup
  \renewcommand\thefootnote{}\footnote{#1}%
  \addtocounter{footnote}{-1}%
  \endgroup
}

\title{Letting Emotions Flow: Success Prediction by Modeling the Flow of Emotions in Books}

\author{Suraj Maharjan$^\star$   Sudipta Kar$^\star$  Manuel Montes-y-Gómez$^\dagger$  \\ {\bf Fabio A. González$^\ddagger$}   {\bf Thamar Solorio$^\star$} \\
$^\star$Department of Computer Science, University of Houston\\
$^\dagger$Instituto Nacional de Astrofısica Optica y Electronica, Puebla, Mexico\\\
$^\ddagger$Systems and Computer Engineering Department, Universidad Nacional de Colombia\\
{\tt \{smaharjan2, skar3\}@uh.edu, \tt solorio@cs.uh.edu }\\
{\tt mmontesg@ccc.inoep.mx}\\
{\tt fagonzalezo@unal.edu.co }}

\date{}
\hypersetup{draft}
\begin{document}
\maketitle
\begin{abstract}
Books have the power to make us feel happiness, sadness, pain, surprise, or sorrow. An author's dexterity in the use of these emotions captivates readers and makes it difficult for them to put the book down. In this paper, we model the flow of emotions over a book using recurrent neural networks and quantify its usefulness in predicting success in books. We obtained the best weighted F1-score of 69\% for predicting books' success in a multitask setting (simultaneously predicting success and genre of books). 
\end{abstract}

\section{Introduction}
Books have the power to evoke a multitude of emotions in their readers. They can make readers laugh at a comic scene, cry at a tragic scene and even feel pity or hate for the characters. Specific patterns of emotion flow within books can compel the reader 
to finish the book, and possibly pursue similar books in the future. 
Like a musical arrangement, the right emotional rhythm can arouse readers, but even a slight variation in the composition might 
turn them away.

\newcite{sunday1981autobiographical} discussed the potential of plotting emotions in stories on the ``Beginning-End" and the ``Ill Fortune-Great Fortune" axes. ~\newcite{emotional_arc_reagan2016}  used mathematical tools like Singular Value Decomposition, agglomerative clustering, and Self Organizing Maps~\cite{Kohonen:2001:SM:558021} to generate basic shapes of stories. They found that stories are dominated by six different shapes. They even correlated these different shapes to the success of books.~\newcite{emotrack} visualized emotion densities across books of different genres. He found that the progression of emotions varies with the genre. For example, there is a stronger progression into darkness in horror stories than in comedy. Likewise, ~\newcite{KAR18.332} showed that movies  having similar flow of emotions across their plot synopses were assigned similar set of tags by the viewers.

\begin{figure}[h]
\centering
\includegraphics[width=\columnwidth]{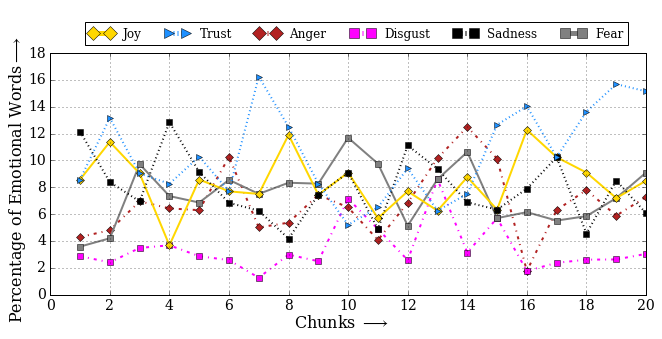}
\caption{ Flow of emotions in \textit{Alice in Wonderland}.}
\label{fig:emotion_flow_plot}
\end{figure}
As an example, in Figure~\ref{fig:emotion_flow_plot}, we draw the flow of emotions across the book: \textit{Alice in Wonderland}. 
The plot shows continuous change in \textit{trust}, \textit{fear}, and \textit{sadness}, which relates to the main character's getting into and out of trouble. These patterns present the emotional arcs of the story. 
Even though they do not reveal the actual plot, they indicate major events happening in the story. 

In this paper, we hypothesize that readers enjoy emotional rhythm and thus modeling emotion flows  will help predicting a book's potential success. In addition, we show that using the entire content of the book yields better results. Considering only a fragment, as done in earlier work that focuses mainly on style \cite{maharjan-EtAl:2017:EACLlong,ganjigunteashok-feng-choi:2013:EMNLP}, disregards important emotional changes. Similar to \newcite{maharjan-EtAl:2017:EACLlong}, we also find that adding genre as an auxiliary task improves success prediction.

\section {Methodology}\blfootnote{The source code and data for this paper can be downloaded from \url{https://github.com/sjmaharjan/emotion_flow}} 
We extract emotion vectors from different chunks of a book and feed them to a recurrent neural network (RNN) to model the sequential flow of emotions. We aggregate the encoded sequences into a single book vector using an attention mechanism.  
Attention models have been successfully used in various Natural Language Processing tasks~\cite{wang-EtAl:2016:EMNLP20163,yang-EtAl:2016:N16-13,NIPS2015_5945,chen-bolton-manning:2016:P16-1,rush-chopra-weston:2015:EMNLP,luong-pham-manning:2015:EMNLP}. 
This final vector, which is emotionally aware, is used for success prediction. \\



\noindent {\bf Representation of Emotions}:
NRC Emotion Lexicons provide $\sim$14K words (Version 0.92) and their binary associations with eight types of elementary emotions (\textit{anger}, \textit{anticipation}, \textit{joy}, \textit{trust}, \textit{disgust}, \textit{sadness}, \textit{surprise,} and \textit{fear}) from the Hourglass of emotions model with polarity (\textit{positive} and \textit{negative}) \cite{COIN:COIN460,mohammad-turney:2010:EMOTION}. These lexicons have been shown to be effective in tracking emotions in literary texts~\cite{emotrack}.\\

\begin{figure}[h]
  \centering
  \includegraphics[width=\columnwidth,keepaspectratio]{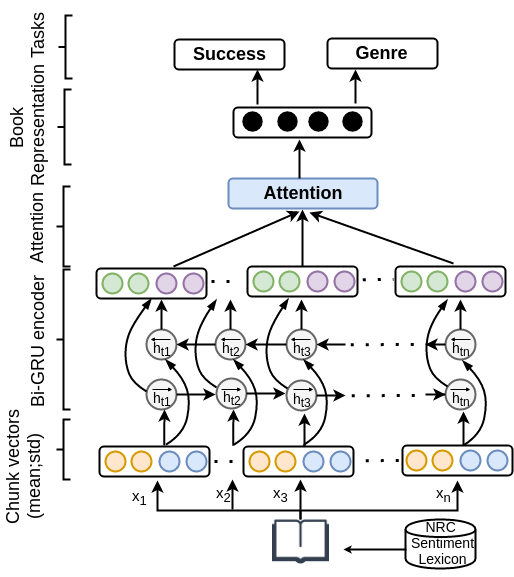}
   \caption{  \label{fig:sentiment_flow}Multitask Emotion Flow Model.}
\end{figure}

\noindent {\bf Inputs}: Let $X$ be a collection of books, where each book $x\in X$ is represented by a sequence of $n$ chunk emotion vectors, $x  = (x_{1}, x_{2}, ..., x_{n})$, where $x_{i}$ is the aggregated emotion vector for chunk $i$, as shown in Figure~\ref{fig:sentiment_flow}. We  divide the book into $n$ different chunks based on the number of sentences.
We then create an emotion vector for each sentence by counting the presence of words of the sentence in each of the ten different types of emotions of the NRC Emotion Lexicons. Thus, the sentence emotion vector has a dimension of 10.
Finally, we aggregate these sentence emotion vectors  into a chunk emotion vector by taking the average and standard deviation of sentence vectors in the chunk. 
Mathematically, the $i$th chunk emotion vector ($x_{i}$) is defined as follows:
\begin{equation}\label{eq:chunk_vector}
	x_{i} = \Bigg[ \frac{\sum_{j=1}^N s_{ij}}{N} ; \sqrt[]{\frac{\sum_{j=1}^N {(s_{ij}-\bar{s_i})}^2}{N}}\Bigg]
\end{equation}
where, $N$ is the total number of sentences, $s_{ij}$ and  $\bar{s_i}$ are the $j$th sentence emotion vector and the mean of the sentence emotion vectors for the $i$th chunk, respectively. The chunk vectors have a dimension of 20 each. The motivation behind using the standard deviation as a feature is to capture the dispersion of emotions within a chunk.\\

\noindent {\bf Model}: 
We then use bidirectional Gated Recurrent Units (GRUs) \cite{DBLP:journals/corr/BahdanauCB14}  to summarize the contextual emotion flow information from both directions. The forward  and backward GRUs will read the sequence from $x_{1}$ to $x_{n}$, and from $x_{n}$ to $x_{1}$, respectively. These operations will compute the forward hidden states  $(\overrightarrow{h_{1}}, \ldots, \overrightarrow{h_{n}})$ and backward hidden states $(\overleftarrow{h_{1}}, \ldots, \overleftarrow{h_{n}})$. 
The annotation for  each chunk $x_{i}$ is obtained by concatenating its forward hidden states $\overrightarrow{h_{i}}$ and its backward hidden states $\overleftarrow{h_{i}}$, i.e. $h_{i}$=$[\overrightarrow{h_{i}}; \overleftarrow{h_{i}}]$. We then learn the relative importance of these hidden states for the classification task. We combine them by taking the weighted sum of all $h_{i}$ and represent the final book vector $r$ using the following equation:

\begin{align}\label{eq:attention}
r &= \sum_{{i}}\alpha_{i} h_{i}\\
\alpha_{i}&=\frac{\exp({score(h_{i})})}{\sum_{{i}'}\exp(score(h_{{i}'}))}\\
score(h_{i})&= v^{T}selu(W_{a}h_{i}+b_a)
\end{align}

where, $\alpha_{i}$ are the weights, $W_a$ is the weight matrix, $b_a$ is the bias, $v$ is the weight vector, and $selu$~\cite{selu} is the nonlinear activation function. Finally, we apply a linear transformation  that maps  the book vector $r$ to the number of classes. In case of the single task (ST) setting, where we only predict  success, we  apply  sigmoid activation to get the final prediction probabilities and compute errors using binary cross entropy loss. Similarly, in the multitask (MT) setting, where we predict both success and  genre~\cite{maharjan-EtAl:2017:EACLlong},  we apply a softmax activation to get the final prediction probabilities for the genre prediction. Here, we add the losses from both tasks, i.e. $L_{total}$=$L_{suc}$ + $L_{gen}$ ($L_{suc}$ and $L_{gen}$ are success and genre tasks' losses, respectively), and then train the network using backpropagation.
%
%
%

\section{Experiments}

\subsection{Dataset}

We experimented with the dataset introduced by~\newcite{maharjan-EtAl:2017:EACLlong}. The dataset consists of 1,003  books  from eight different genres collected from Project Gutenberg\footnote{\footnotesize{https://www.gutenberg.org/}}. The authors considered only those books that were at least reviewed by ten reviewers. They  categorized these books into two \textcolor{black}{ classes}, \textit{Successful} (654 books) and \textit{Unsuccessful} (349 books), based on the average rating for the books in Goodreads\footnote{\footnotesize{https://www.goodreads.com/}} website. They considered only the first 1K sentences from each book. 

\subsection{Baselines}
We compare our proposed methods with the following baselines:

\noindent {\bf Majority Class}: The majority class in training data is \textit{success}. This baseline obtains a weighted F1-score of 0.506 for all the test instances. 

\noindent {\bf SentiWordNet+SVM}: ~\newcite{maharjan-EtAl:2017:EACLlong} used SentiWordNet~\cite{baccianella2010sentiwordnet} to compute the sentiment features along with counts of different Part of Speech (PoS) tags for every 50 consecutive sentences (20 chunks from 1K sentences) and used an SVM classifier.

\noindent {\bf NRC+SVM}:  We concatenate the chunk emotion vectors ($x_{i}$) created using the NRC lexicons and feed them to the SVM classifier. We experiment by varying the number of book chunks.

These baseline methods do not incorporate the sequential flow of emotions across the book and treat each feature independently of each other. 
\subsection{Experimental Setup}
 We experimented with the same random stratified splits of 70:30 training  to test ratio as used by~\newcite{maharjan-EtAl:2017:EACLlong}. We use the SVM algorithm for the baselines and RNN for our proposed emotion flow method. We tuned the \textit{C} hyperparameter of the SVM classifier by performing grid search on the values (1e\{-4,...,4\}), using three fold cross validation on the training split. For the experiments with RNNs, we first took a random stratified split of 20\% from the training data as validation set. We then tuned the RNN hyperparameters by running 20 different experiments with a random selection of different values for  the  hyperparameters. We tuned the weight initialization (Glorot Uniform~\cite{pmlr-v9-glorot10a}, LeCun Uniform~\cite{LeCun:1998:EB:645754.668382}), learning rate with Adam~\cite{adam2015}  \{1e-4,\ldots,1e-1\}, dropout rates \{0.2,0.4,0.5\}, attention and recurrent units \{32, 64\}, and batch-size \{1, 4, 8\} with early stopping criteria.

\section{Results}

\begin{table}[!h]
\centering
\resizebox{\columnwidth}{!}{%
\begin{tabular}{|l|r|r|r|r|r|}
\hline
{\bf Book Content}       & \multicolumn{1}{l|}{}       & \multicolumn{2}{c|}{\bf 1000 sents} & \multicolumn{2}{c|}{\bf All}        \\ \hline
{\bf Methods}            & \multicolumn{1}{c|}{\bf Chunks} & \multicolumn{1}{c|}{\bf ST}    & \multicolumn{1}{c|}{\bf MT}    & \multicolumn{1}{c|}{\bf ST}    & \multicolumn{1}{c|}{\bf MT}    \\ \hline\hline
Majority Class     & -   & 0.506  & 0.506  & 0.506  & 0.506  \\ 
SentiWordNet + SVM & 20  & 0.582  & 0.610  & -  & -  \\ \hline \hline
NRC + SVM           & 10  & 0.526  & 0.597  & 0.541  & 0.641  \\ 
NRC + SVM           & 20  & 0.537  & 0.590  & 0.577  & 0.604  \\ 
NRC + SVM           & 30  & 0.587  & 0.576  & 0.595  & 0.600  \\ 
NRC + SVM           & 50  & 0.611  & 0.586  & 0.597  & 0.636 \\ \hline\hline
Emotion Flow      & 10  & 0.632  & 0.643  & 0.650  & 0.660   \\ 
Emotion Flow      & 20  & 0.612  & 0.639  & 0.640  & 0.668   \\ 
Emotion Flow      & 30  & 0.630  & 0.657  & 0.662  & 0.677  \\ 
Emotion Flow      & 50  & 0.656  & 0.666  & 0.674  & \textbf{0.690}*  \\ \hline
\end{tabular}
 }
\caption{ \label{table:sf_results} Weighted F1-scores for  success classification in single task (ST) and multi task (MT) settings with varying chunk sequences when using all the book or only the first 1K sentences. $*p<0.05$ (McNemar significance test between Emotion Flow (chunks  50, MT, All) and NRC+SVM (chunk 10, MT, All))}
\end{table}

%
%
%


Table~\ref{table:sf_results} presents the results. Our proposed method performs better than different baseline methods and obtains the highest weighted F1-score of 0.690. The results highlight the importance of taking into account the sequential flow of emotions across books to predict how much readers will like a book. We obtain better performance when we use an RNN to feed the sequences of emotion chunk vectors. The performance decreases with the SVM classifier, which discards this sequential information by treating each feature independently of each other. 
Moreover, increasing the granularity of the emotions by increasing the number of chunks seems to be helpful for success prediction. However, we see a slight decrease in performance beyond 50 chunks (weighted F1 score of 0.662 and 0.664 for 60 and 100 chunks, respectively).

The results also show that the MT setting is beneficial over the ST setting, whether we consider the first 1K sentences or the entire book. This finding is akin to \newcite{maharjan-EtAl:2017:EACLlong}. Similar to them, we suspect the auxiliary task of genre classification  is acting as a regularizer.

Considering only the first 1K sentences of books may miss out important details, especially when the only input to the model is the distribution of emotions. It is necessary to include information from  later chapters and  climax of the story as they gradually reveal the answers to the suspense, the events, and the emotional ups and downs in characters that build up through the course of the book. Accordingly, our results show that it is important to consider emotions from the entire book rather than from just the first 1K sentences. 


\section{Attention Analysis}
\pgfplotstableread
{data/cmp_gru_att.dat}
{\gruattntable}

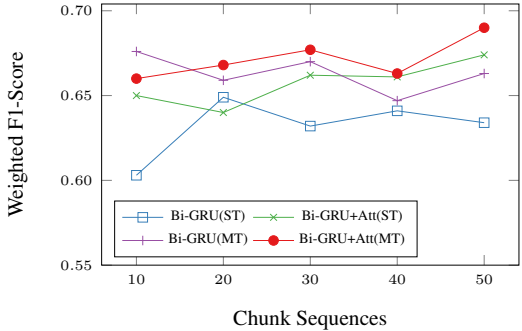
\begin{figure}[ht]
\centering
\resizebox{0.9\columnwidth}{!}{%
\begin{tikzpicture}
\begin{axis}[
    ytick = {0.50,0.55,0.60,0.65,0.70,0.80,0.90,1.00}, 
    y tick label style={
        /pgf/number format/.cd,
            fixed,
            fixed zerofill,
            precision=2,
        /tikz/.cd
    },
    xtick={10,20,30,40,50,60},
    ymin=0.55,
    width=\columnwidth,
    legend style={font=\tiny,legend pos=south west},
    legend columns=2, 
	height = 0.34\textwidth,
    xlabel =Chunk Sequences,
    ylabel = Weighted F1-Score,
    label style={font=\small},
	tick label style={font=\tiny},
	cycle list name=sf_colors]
    \addplot table[x=Chunks, y=Bi-GRU(ST)] from \gruattntable;
    \addplot table[x=Chunks, y=Bi-GRU+Att(ST)] from \gruattntable;
    \addplot table[x=Chunks, y=Bi-GRU(MT)] from \gruattntable;
    \addplot table[x=Chunks, y=Bi-GRU+Att(MT)] from \gruattntable;
    \legend{Bi-GRU(ST),Bi-GRU+Att(ST),Bi-GRU(MT),Bi-GRU+Att(MT)}
    \end{axis}
    \end{tikzpicture}
 }
\caption{ \label{fig:cmp_gru_att} \textcolor{black}{Comparison of the Emotion Flow  with and without attention mechanism for different chunk sequences.} }
%
%
\end{figure}
From Figure~\ref{fig:cmp_gru_att}, we see that using the attention mechanism to aggregate vectors is better than just concatenating the final forward and backward hidden states to represent the book in both ST and MT settings.
We also observe that the multitask approach performs better than the singe task one regardless of the number of chunks and the use of attention.


\begin{figure}[!h]
  \centering
  \includegraphics[width=\columnwidth, height=4.5cm]{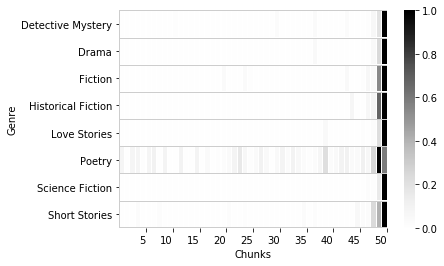}
   \caption{  \label{fig:attention_viz}Attention weights visualization per genre.}
\end{figure}

Figure~\ref{fig:attention_viz} plots the heatmap of the average attention weights for test books grouped by their genre. The model has learned that the last two to three chunks that represent the climax, are most important for predicting success. Since this is a bidirectional RNN model, hidden representations for each chunk carry information from the whole book.
Thus, using only the last chunks will probably result in lower performance.
Also, the weights visualization \textcolor{black}{shows} an interesting pattern for \textit{Poetry}. In \textit{Poetry}, emotions are distributed across the different regions. 
This may be due to sudden important events or abrupt change in emotions. For \textit{Short stories}, initial chunks also receive some weights, suggesting the importance of the premise.
%
%

\section{ Emotion Analysis}
\pgfplotstableread[col sep=tab]
{data/feature_ranking.dat}
{\featurerankdata}

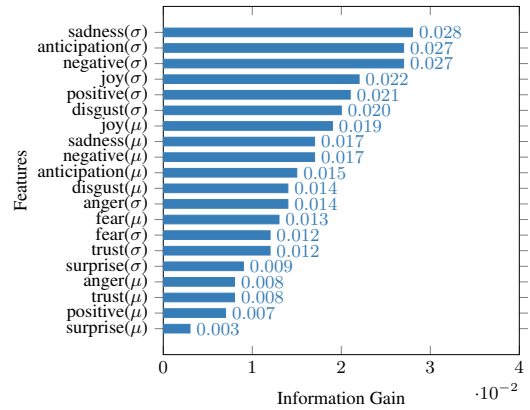
\begin{figure}[h]
\centering
\small
\resizebox{0.9\columnwidth}{!}{%
\begin{tikzpicture}
  \begin{axis}[
      xbar,
      y=-0.27cm,
      bar width=0.15cm,
      enlarge y limits={abs=0.45cm},      
      width=\columnwidth,
      xmin=0.0, xmax=0.04,
      yticklabels from table = {\featurerankdata}{Feature},
      nodes near coords,
      nodes near coords style={/pgf/number format/.cd,fixed zerofill,fixed,precision=3},
      nodes near coords align={horizontal},
      ytick={0,...,19}, 
      cycle list name=bar_colors,
      xlabel={Information Gain},
      ylabel={Features},
      y label style={at={(-0.2,0.5)}}
  ]
  \addplot table[y expr=\coordindex,x=Infogain]{\featurerankdata};
  \end{axis}
\end{tikzpicture}
}

\caption{  \label{fig:feature_ranking}  Feature ranking with information gain.}
\end{figure}
\noindent{\bf Climax Emotions}: Since the last chunk is assigned more weights than other chunks, we used information gain to rank features of that chunk.
From Figure~\ref{fig:feature_ranking}, we see that features capturing the variation of different emotions are ranked higher than features capturing the average scores. This suggests that readers tend to enjoy emotional ups and downs portrayed in books, making the standard deviation features more important than the average features for the same emotions.

\begin{table*}[!h]
\centering
\resizebox{\textwidth}{!}{%
\begin{tabular}{|l|rr|rr|rr|rr|rr|rr|rr|rr|}
\hline
             & \multicolumn{2}{c|}{\textbf{Anger}}                   & \multicolumn{2}{c|}{\textbf{Anticipation}}            & \multicolumn{2}{c|}{\textbf{Disgust}}                 & \multicolumn{2}{c|}{\textbf{Fear}} & \multicolumn{2}{c|}{\textbf{Joy}}                 & \multicolumn{2}{c|}{\textbf{Sadness}}                 & \multicolumn{2}{c|}{\textbf{Surprise}}                & \multicolumn{2}{c|}{\textbf{Trust}}                   \\ \hline
Dataset      & \multicolumn{1}{c}{$\mu$} & \multicolumn{1}{c|}{$\sigma$} & \multicolumn{1}{c}{$\mu$} & \multicolumn{1}{c|}{$\sigma$} & \multicolumn{1}{c}{$\mu$} & \multicolumn{1}{c|}{$\sigma$} & \multicolumn{1}{c}{$\mu$} & \multicolumn{1}{c|}{$\sigma$} & \multicolumn{1}{c}{$\mu$} & \multicolumn{1}{c|}{$\sigma$} & \multicolumn{1}{c}{$\mu$} & \multicolumn{1}{c|}{$\sigma$} & \multicolumn{1}{c}{$\mu$} & \multicolumn{1}{c|}{$\sigma$} & \multicolumn{1}{c}{$\mu$} & \multicolumn{1}{c|}{$\sigma$}  \\ \hline
Corpus       & 0.248   & 0.249      & 0.414   & 0.372      & 0.179   & 0.200      & 0.340   & 0.327      & 0.399   & 0.431            & 0.323   & 0.309      & 0.225   & 0.191      & 0.492   & 0.441      \\ 
Successful   & 0.270   & 0.263      & 0.447   & 0.374      & 0.194   & 0.207      & 0.377   & 0.353      & 0.435   & 0.416           & 0.358   & 0.334      & 0.236   & 0.207      & 0.517   & 0.427      \\ 
Unsuccessful & 0.207   & 0.214      & 0.351   & 0.359      & 0.153   & 0.183      & 0.270   & 0.258      & 0.331   & 0.451         & 0.258   & 0.243      & 0.205   & 0.155      & 0.445   & 0.463      \\ \hline
\end{tabular}
}
\caption{ \label{table:avg_scores}Mean ($\mu$) and standard  deviation ($\sigma$) for eight type of emotions  for the last chunk.}
\end{table*}

Table~\ref{table:avg_scores} shows the mean ($\mu$) and standard deviation ($\sigma$) for different emotions extracted for all the data, and further categorized by \textit{Successful} and \textit{Unsuccessful} label from the last chunk. We see that authors generally end books with higher rates of positive words ($\mu=0.888$) than negative words ($\mu=0.599$) and the difference is significant ($p<0.001$). Similarly the means for \textit{anticipation}, \textit{joy}, \textit{trust}, and \textit{fear} are higher than for \textit{sadness}, \textit{surprise},  \textit{anger}, and \textit{disgust}. This further validates that authors prefer happy ending. Moving on to \textit{Successful} and \textit{Unsuccessful} categories, we see that the means for \textit{Successful} books are higher than \textit{Unsuccessful} books for \textit{anger}, \textit{anticipation}, \textit{disgust}, \textit{fear}, \textit{joy}, and \textit{sadness} (highly significant, $p<0.001$). We observe the same pattern for \textit{trust}, and \textit{surprise}, although the $p$ value is only $p<0.02$ in this case. Moreover, the standard deviations for all emotions are significantly different across the two categories ($p<0.001$). Thus, emotion concentration ($\mu$) and variation ($\sigma$) for \textit{Successful} books are higher than for \textit{Unsuccessful} books for all emotions in the NRC lexicon. 



\pgfplotstableread
{data/centriods_50_joy_T.tsv}
{\centroid}

  



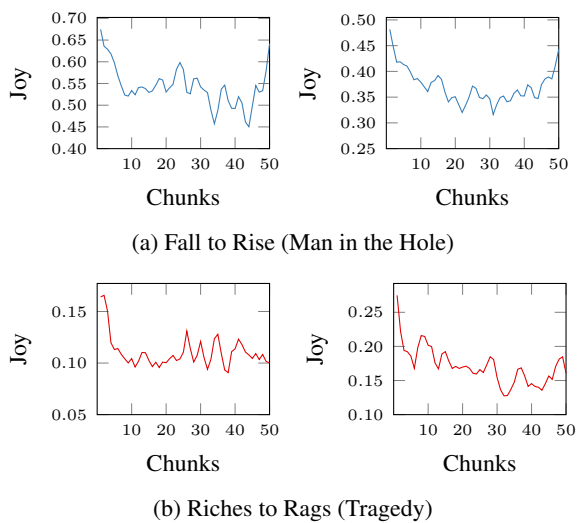
\begin{figure}[!h]
\centering
\makebox[\columnwidth][c]{%
    \begin{subfigure}[t]{\columnwidth}
        \centering
        \begin{tikzpicture}
\begin{axis}[
    ytick = {0.40,0.45,0.50,0.55,0.60,0.65,0.70}, 
    y tick label style={
        /pgf/number format/.cd,
            fixed,
            fixed zerofill,
            precision=2,
        /tikz/.cd
    },
    xtick={10,20,30,40,50},
    ymin=0.40,
    xmax=50,
    xmin=0,
    width=0.5\columnwidth,
    xlabel =Chunks,
    ylabel =Joy,
    ylabel near ticks,
    xlabel near ticks,
    label style={font=\footnotesize},
	tick label style={font=\tiny},
	cycle list name=centroid_colors]
    \addplot table[x=Chunks, y=Cluster_37] from \centroid;
    \end{axis}
    \end{tikzpicture}\hfill
\begin{tikzpicture}
\begin{axis}[
    ytick = {0.25,0.30,0.35,0.40,0.45,0.50,0.55,0.60}, 
    y tick label style={
        /pgf/number format/.cd,
            fixed,
            fixed zerofill,
            precision=2,
        /tikz/.cd
    },
    xtick={10,20,30,40,50},
    ymin=0.25,
    xmax=50,
    xmin=0,
    width=0.5\columnwidth,
    xlabel =Chunks,
    ylabel =Joy,
     ylabel near ticks,
    xlabel near ticks,
    label style={font=\footnotesize},
	tick label style={font=\tiny},
	cycle list name=centroid_colors]    
     \addplot table[x=Chunks, y=Cluster_54] from \centroid;   
    \end{axis}
    \end{tikzpicture}
    \normalsize{
    \caption{Fall to Rise (Man in the Hole)}}
    \end{subfigure}
}\\
\vspace{0.3cm}
\makebox[\columnwidth][c]{%
\begin{subfigure}[t]{\columnwidth}
\centering
\begin{tikzpicture}
\begin{axis}[
    ytick = {0.05,0.10,0.15,0.20}, 
    y tick label style={
        /pgf/number format/.cd,
            fixed,
            fixed zerofill,
            precision=2,
        /tikz/.cd
    },
    xtick={10,20,30,40,50},
    ymin=0.05,
    xmax=50,
    xmin=0,
    width=0.5\columnwidth,
    xlabel =Chunks,
    ylabel =Joy,
    ylabel near ticks,
    xlabel near ticks,
    label style={font=\footnotesize},
	tick label style={font=\tiny},
	cycle list name=centroid_colors1]
    \addplot table[x=Chunks, y=Cluster_83] from \centroid;
    \end{axis}
    \end{tikzpicture}\hfill
\begin{tikzpicture}
\begin{axis}[
    ytick = {0.10,0.15,0.20,0.25,0.30,0.35}, 
    y tick label style={
        /pgf/number format/.cd,
            fixed,
            fixed zerofill,
            precision=2,
        /tikz/.cd
    },
    xtick={10,20,30,40,50},
    ymin=0.10,
    xmax=50,
    xmin=0,
    width=0.5\columnwidth,
    xlabel =Chunks,
    ylabel =Joy,
    ylabel near ticks,
    xlabel near ticks,
    label style={font=\footnotesize},
	tick label style={font=\tiny},
	cycle list name=centroid_colors1]    
     \addplot table[x=Chunks, y=Cluster_90] from \centroid;   
    \end{axis}
    \end{tikzpicture}
    \caption{Riches to Rags (Tragedy)}
    \end{subfigure}
 }

\caption{ \label{fig:emotion_shapes} {Emotion flow for four cluster centroids in the dataset. The two curves on top match the ``Fall to Rise'' shape and the two at the bottom match the ``Tragedy'' one defined in \newcite{emotional_arc_reagan2016}.}}
\end{figure}

\noindent{\bf Emotion Shapes}: We visualize the prominent emotion flow shapes in the dataset using K-means clustering algorithm. We took the average \textit{joy} across 50 chunks for all books and clustered them into 100 different clusters. We then plotted the smoothed centroid of clusters having $\geq$ 20 books. We found two distinct shapes ( ``Man in the hole" (fall to rise) and “Tragedy” or ``Riches to rags" (fall)). Figure~\ref{fig:emotion_shapes} shows such centroid plots. The plot also shows that the ``Tragedy" shapes have an overall lower value of joy than the ``Man in the hole" shapes. Upon analyzing the distribution of \textit{Successful} and \textit{Unsuccessful} books within these shapes, we found that the ``Man in the hole" shapes have a higher number of successful books whereas, the ``Tragedy” shapes have the opposite. 


\section{Conclusions and Future Work}
In this paper, we showed that 
modeling emotions as a flow, by capturing the emotional content at different stages, improves prediction accuracy.
We learned that most of the attention weights are given to the last fragment in all genres, except for \textit{Poetry} where  other fragments seem to be relevant as well. We also showed empirically that adding an attention mechanism is better than just considering the last forward and backward hidden states from the RNN. We found two distinct emotion flow shapes and found that the clusters with ``Tragedy” shape had more unsuccessful books than successful ones. In future work, we will be exploring how we can use these flows of emotions to detect important events that result in suspenseful scenes. Also, we will be applying hierarchical methods that take in the logical grouping of books (sequence of paragraphs to form a chapter and sequence of chapters to form a book) to build books' emotional representations. 

\section*{Acknowledgments}
We would like to thank the National Science Foundation for funding this work under award 1462141. We are also grateful to the anonymous reviewers for reviewing the paper and providing helpful comments and suggestions.


\bibliography{naaclhlt2018}
\bibliographystyle{acl_natbib}
\end{document}